\documentclass[nohyperref]{article}

\usepackage{microtype}
\usepackage{graphicx}
\usepackage{subfigure}
\usepackage{booktabs} 

\usepackage{hyperref}
\usepackage{pifont}

\usepackage[accepted]{icml2022}

\usepackage{amsmath}
\usepackage{amssymb}
\usepackage{mathtools}
\usepackage{amsthm}

\usepackage[capitalize,noabbrev]{cleveref}

\theoremstyle{plain}

\theoremstyle{definition}

\theoremstyle{remark}

\usepackage[textsize=tiny]{todonotes}

\icmltitlerunning{Universality of Winning Tickets: A Renormalization Group Perspective}

\begin{document}

\twocolumn[
\icmltitle{Universality of Winning Tickets: A Renormalization Group Perspective}




\begin{icmlauthorlist}
\icmlauthor{William T. Redman}{UCSB}
\icmlauthor{Tianlong Chen}{UTA}
\icmlauthor{Zhangyang Wang}{UTA}
\icmlauthor{Akshunna S. Dogra}{ICL,EPSRC}
\end{icmlauthorlist}

\icmlaffiliation{UCSB}{Interdepartmental Graduate Program in Dynamical Neuroscience, University of California, Santa Barbara.}
\icmlaffiliation{UTA}{Department of Electrical and Computer Engineering, University of Texas at Austin.}
\icmlaffiliation{ICL}{Department of Mathematics, Imperial College London.}
\icmlaffiliation{EPSRC}{EPSRC CDT in Mathematics of Random Systems: Analysis, Modelling and Simulation}

\icmlcorrespondingauthor{William T. Redman}{wredman@ucsb.edu}

\icmlkeywords{Lottery ticket hypothesis, transfer, renormalization group theory}

\vskip 0.3in
]



\printAffiliationsAndNotice{}  

\begin{abstract}
Foundational work on the Lottery Ticket Hypothesis has suggested an exciting corollary: winning tickets found in the context of one task can be transferred to similar tasks, possibly even across different architectures. This has generated broad interest, but methods to study this universality are lacking. We make use of renormalization group theory, a powerful tool from theoretical physics, to address this need. We find that iterative magnitude pruning, the principal algorithm used for discovering winning tickets, is a renormalization group scheme, and can be viewed as inducing a flow in parameter space. We demonstrate that ResNet-50 models with transferable winning tickets have flows with common properties, as would be expected from the theory. Similar observations are made for BERT models, with evidence that their flows are near fixed points. Additionally, we leverage our framework to study winning tickets transferred across ResNet architectures, observing that smaller models have flows with more uniform properties than larger models, complicating transfer between them. 
\end{abstract}

\section{Introduction}
\label{Introduction}

The lottery ticket hypothesis (LTH) for deep neural networks (DNNs) proposes that DNNs contain sparse subnetworks that can be trained in isolation and reach performance that is equal to, or better than, that of the full DNN in the same number of training iterations \citep{fra19, fra20}. These subnetworks are called winning lottery tickets. The LTH has provided paradigm shifting insight into the success of dense DNNs, suggesting that the emergence of winning tickets, with increasing DNN size, plays a key role, reminiscent of the maxim ``more is different'' \citep{and72}. 

In recent years, researchers have found an intriguing corollary: winning tickets found in the context of one task can be transferred to related tasks \citep{desai2019shifts, meh19, mor19, soe19, che20, goh20, che21, sab21}, possibly even across architectures \citep{chen2021ELTH}. Their existence has recently been proven \citep{burkholz2022on}, further supporting the idea that they are a general phenomenon. In addition to having applications of practical interest, these results imply that winning tickets can be used to study how tasks and architectures are ``similar''. However, there currently exist few tools with which to study this universality, and there exists no way to know, without directly performing transfer experiments, which previously studied tasks a given winning ticket can be transferred to. These are, in part, due to the fact that there is a general lack of theoretical work on iterative magnitude pruning (IMP) [although see \citet{tanaka2020pruning} and \citet{ele21}], the most commonly used method for finding winning tickets.

This is in striking analogy to the state of statistical physics in the early-to-mid--20$^{\text{th}}$ century. Empirical evidence suggested that disparate systems, governed by seemingly different underlying physics, exhibited the \textit{same}, universal properties near their phase transitions. While heuristic methods provided insight \citep{kad66}, a full theory from first principles was not realized until the development of the renormalization group (RG) \citep{wil71A, wil71B, wil75}.

RG theory has not only provided a framework for explaining universal behavior near phase transitions, but also a scheme for grouping systems by that behavior. This classification, introduced via the notion of universality classes, has allowed for a detailed understanding of materials \citep{tou97,win97, dur00, bon10}. In addition, it provides knowledge as to what previously classified materials a newly studied substance behaves like (see Appendix \ref{RG theory in physics} for additional background on RG theory in physics).

The analogy between universality in RG and LTH theories again emerges when considering the recent work of \citet{ros21}. It was found that, when the density (i.e. the percentage of parameters remaining) of a DNN being pruned via IMP is within a certain range, $d_L < d < d_C$, the DNN's error scales according to a power-law,
\begin{equation}
    \label{eq:pruning scaling law}
    e \sim (d_C - d)^{-\gamma} = \Delta d^{-\gamma}.
\end{equation}
Power-law scaling is well known to emerge in critical phenomena that the RG is used to study. For instance, when the temperature of a classical spin system (e.g. two-dimensional Ising model) is near, but below the critical temperature ($t < t_C$), many observables (e.g. magnetization) exhibit power-law scaling,
\begin{equation}
    \label{eq:magnetization scaling law}
    m \sim (t_C - t)^{-\beta} = \Delta t ^{-\beta}.
\end{equation}

\begin{figure*}[t]
    \centering
    \includegraphics[width = 0.75\textwidth]{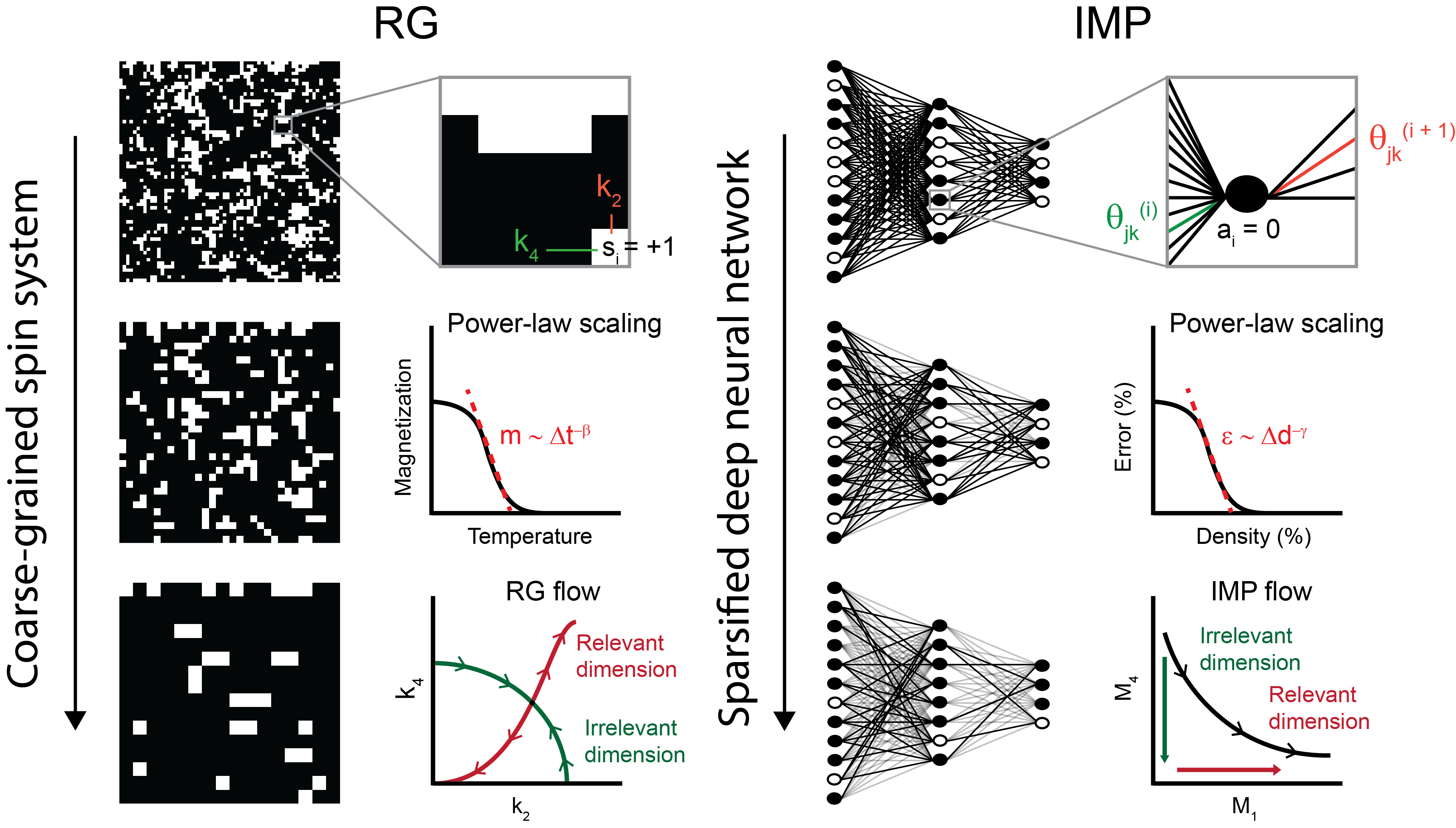}
    \caption{\textbf{Summary of the similarities between the RG and IMP.} Both the RG and IMP are applied iteratively to coarse-grain systems, revealing ``relevant'' features. Certain observables are known to have regimes where they follow power-law scaling. In the case of the RG, this scaling, and its universality, are associated with properties of the flow that the RG induces in the space of coupling constants (see Sec. \ref{RG Theory RG and LTH}). The nature of the flow that IMP induces has not been previously studied, but none-the-less exists (see Sec. \ref{Results Eigenfunctions}). Table 1 outlines the analogous quantities of each theory.}
    \label{fig:RG IMP}
\end{figure*}

These similarities hint at a more fundamental connection between the universality present in winning tickets and in physical systems (see Fig. \ref{fig:RG IMP} and Table \ref{tab:RG IMP equiv}). Indeed, we show that \textit{IMP is an RG scheme} (i.e. IMP satisfies the properties required to be considered an RG operator). By analyzing large-scale winning ticket transfer experiments \citep{che20, che21}, we find that models which allow for successful transfer have similar behavior in their IMP flow (i.e. the way in which the parameters of the DNN change after each round of IMP), in agreement with the theory. In addition, we observe that the IMP flow is static for tickets found on pre-trained transformers (BERT), whereas it is dynamic for winning tickets found on pre-trained and randomly initialized convolutional neural networks (ResNet-50). We further leverage our RG framework to interpret recent results on transferring winning tickets across differing architectures \citep{chen2021ELTH}, finding that the residual blocks of smaller ResNet architectures are more uniform in their sensitivity to pruning than those of larger ResNet architectures. This is consistent with the experimental result that tickets found on smaller architectures did not transfer as well to larger ones. 

Taken together, we believe our work shows that RG theory is not only an appropriate language with which to study IMP (given that IMP \textit{is} an RG scheme), but is also a useful one that provides a new perspective on experimental results. We hope that our work encourages further collaboration between those studying machine learning  and those studying the RG and statistical physics.

\begin{table}[!htb]
\centering
\label{tab:RG IMP equiv}
\caption{Analogous quantities in RG and IMP theory.}
\begin{tabular}{@{}cc}
\toprule
\multicolumn{1}{c}{\bf RG} &\multicolumn{1}{c}{\bf IMP}\\
\midrule
Spins ($s_i$) & Unit activations ($a_i$) \\
Coupling constants ($k_i$) & Parameters ($\theta_i$)\\
Hamiltonian ($\mathcal{H}[\textbf{s}, \textbf{k}]$) & Loss function ($\mathcal{L}[\textbf{a}, {\boldsymbol{\theta}}]$)  \\
\bottomrule
\end{tabular}
\end{table}

\section{Related Work}

\subsection{Transfer of Winning Tickets}
\label{Related Work Transfer}
The idea that winning tickets found on one task can be successfully transferred to other tasks was first proposed by \citet{fra19} in their original description of the LTH. Ensuing work confirmed this, showing that winning tickets are able to perform well on a variety of tasks beyond the one that they were originally discovered on \citep{desai2019shifts, meh19, mor19, soe19, goh20, sab21}. This has suggested that such tasks share similar underlying properties, in-line with intuition that they are ``related'', and that winning tickets are able to perform universal computations. Winning tickets from DNNs pre-trained on large, complex data sets have shown transferability to a number of ``downstream'' tasks \citep{che20, che21}, further supporting the idea that the solution to a challenging task contains solutions to simpler tasks. Finally, a method for transforming winning tickets to fit different architectures has recently been developed \citep{chen2021ELTH}, highlighting the existence of common structure in winning tickets across families of DNN architectures. 

To-date, the primary approach for studying winning ticket transfer, on a fixed architecture, is to first find winning tickets for tasks $A$ and $B$. The winning ticket found on task $A$ is then applied to task $B$ (and vice versa), and if the performance is similar to the original winning ticket, then the two tasks are said to admit transfer. While straightforward, such an approach makes it computationally expensive to assess the transferability of many tasks (e.g. tasks $A, B, C, ...$), as each combination has to be assessed. Additionally, the components of the winning tickets that are common across tasks are not readily identified. 

In this work, we develop an approach based on RG theory to quantitatively characterize winning tickets found on different tasks, and different architectures. This characterization is succinct, and winning tickets with the same properties are transferable. Additionally, it identifies some aspects of transferable winning tickets that are common across tasks and architectures. 

\subsection{Renormalization Group Theory and Machine Learning}
\label{Related Work RG}

The success of RG theory in studying emergent phenomena in physical systems has led to interest in applying it to machine learning. For instance, Restricted Boltzmann Machines (RBMs) have been proposed to have an exact mapping to the RG \citep{meh14} [although see comments by \citet{lin2017does} and the response by \citet{schwab2016comment}]. RBMs trained on data generated from Ising models having been found to perform RG-like coarse graining operations \citep{meh14, koch2020deep} [although potentially with different properties from the RG \citep{iso2018scale, koch2020deep}]. Additionally, an RG theory has recently been developed for studying and classifying the output of trained DNNs \citep{rob21}. Lastly, theoretical work has been performed to bridge the language of RG theory to perspectives that are more familiar to machine learning practitioners, such as information theory \citep{koch-Janusz2018mutualinfo, gordon2021relevance} and principal component analysis \citep{bradde2017pca}. 

To our knowledge, RG theory has not yet been applied to the study of sparse machine learning. As we show, there is a direct connection between IMP and the RG (Fig. \ref{fig:RG IMP} and Table \ref{tab:RG IMP equiv}), implying that RG theory is an appropriate language with which to study IMP.

\section{The Renormalization Group and Iterative Magnitude Pruning}
\label{RG Theory}

The RG operator, $\mathcal{R}$, can be viewed as a method for ``coarse-graining''. That is, each application of $\mathcal{R}$ replaces local degrees of freedom with a composite of their values\footnote{Note that here we consider the block spin, or real space, RG. The momentum space RG has a different, but related, interpretation.}. An example of this for the two-dimensional (2D) Ising model is shown in the left-hand column of Fig. \ref{fig:RG IMP}, where neighborhoods of four spins are iteratively replaced by their mode. 

The formal way to study $\mathcal{R}$ is to consider its action on a given Hamiltonian (i.e. energy function). For classical spin systems (e.g. 2D Ising model), $\mathcal{H}$ has the general form 
\begin{equation}
    \label{eq: spin hamiltonian}
    \mathcal{H}(\textbf{s}, \textbf{k}) = -\sum_i k_1 s_i - \sum_{\langle i, j\rangle} k_2 s_is_j - ... ,
\end{equation}
where the $s_i$ are the spins of the system (e.g. $s_i \in \{-1, +1\}$), $\langle \cdot, \cdot \rangle$ represents sites on the lattice that are nearest neighbors, and the $k_i$ are the strengths of the different coupling constants (e.g. $k_2$ is the strength of the nearest neighbor coupling). 

Due to the fact that coarse-graining amalgamates spins, the spin system resulting from an application of $\mathcal{R}$ can be viewed as equivalent to the original, but with a new set of coupling constants. Therefore, the coarse-grained system has a different Hamiltonian, which is given by 
\begin{equation}
    \label{eq: RG equation}
    \mathcal{R} \mathcal{H}(\textbf{s}, \textbf{k}) = \mathcal{H}(\textbf{s}', \mathcal{T} \textbf{k}) = \mathcal{H}(\textbf{s}', \textbf{k}'),
\end{equation}
where $\textbf{s}'$ is the new set of spins and $\textbf{k}'$ is the new set of couplings determined by the operator $\mathcal{T}: \mathbb{R}^{K} \rightarrow \mathbb{R}^K$. Here $K$ is the maximum number of couplings considered, usually introduced to keep the considered operators finite.

$\mathcal{R}$ can be applied iteratively, defining a flow through the function space of Hamiltonians via Eq. \ref{eq: RG equation}, with an associated flow in the space of coupling constants. The latter is commonly referred to as the RG flow. While $\mathcal{T}$ is often a complicated, non-linear function, it can be linearized near fixed points of the flow \citep{gol05}. The flow will grow in the direction of the eigenvectors $\textbf{v}_i$ with eigenvalues $\lambda_i > 1$ and will shrink along the direction of the eigenvectors $\textbf{v}_i$ with $\lambda_i < 1$. These eigenvectors are called \textbf{relevant} and \textbf{irrelevant}, respectively, and are defined in relation to a given fixed point. A schematic of these, and the general RG flow, is presented in Fig. \ref{fig:RG IMP}. By examining the effect of applying $\mathcal{R}$ to a given system, and its flow in the space of coupling constants, it is possible to find which components of the system are necessary for certain macroscopic behavior (i.e those that are relevant), and which are not. Systems belonging to a particular universality class have the same relevant directions.

Interestingly, many of the properties discussed above can be analogously found in the context of IMP. To see this, consider a DNN with loss function $\mathcal{L}(\textbf{a}, \boldsymbol{\theta})$, where $\textbf{a}$ is the unit activations and $\boldsymbol{\theta}$ is the DNN parameters. Let $\mathcal{I}$ represent a single application of the IMP process. The DNN pruned via $\mathcal{I}$ is given by the relation
\begin{equation}
    \label{eq:IMP eq}
    \mathcal{I} \mathcal{L}(\textbf{a}, \boldsymbol{\theta}) =  \mathcal{L}(\textbf{a}', \mathcal{T}\boldsymbol{\theta}) = \mathcal{L}(\textbf{a}', \boldsymbol{\theta}'),
\end{equation}
where the new set of parameters, $\boldsymbol{\theta}'$, are given by an operator $\mathcal{T}$. This new set of parameters leads to a new, coarse-grained set of activations, $\textbf{a}'$. The similarities between Eqs. \ref{eq: RG equation} and \ref{eq:IMP eq} are striking. 

We note here that, in the case of IMP, $\mathcal{T}$ is the composition of a masking operator, $\mathcal{M}$, and a refining operator $\mathcal{F}$. That is, $\mathcal{T} = \mathcal{F} \circ \mathcal{M}: \mathbb{R}^N \rightarrow \mathbb{R}^N$, where $N$ is the number of parameters in the DNN. $\mathcal{M}$ is defined via the pruning procedure implemented (e.g. magnitude pruning). While we are considering IMP here, because of its connection to the LTH, there are many other pruning procedures [e.g. Hessian based pruning \citep{lecun89}], all of which have their own $\mathcal{M}$. Similarly, $\mathcal{F}$ is defined via the refinement procedure used, making it dependent on the choice of optimizer and whether or not the DNN parameters are left alone or ``rewound'' to their value at a previous point during training \citep{fra20, ren20}. 

For $n$ rounds of $\mathcal{I}$, the resulting DNN is given by
\begin{equation}
    \label{eq:IMP flow}
    \mathcal{I}^n \mathcal{L}(\textbf{a}^{(0)}, \boldsymbol{\theta}^{(0)}) = \mathcal{L}(\textbf{a}^{(n-1)}, \mathcal{T}^n\boldsymbol{\theta}^{(0)}) =  \mathcal{L}(\textbf{a}^{(n-1)}, \boldsymbol{\theta}^{(n-1)}).
\end{equation}
This defines a trajectory in parameter space, $\boldsymbol{\theta}^{(0)} \rightarrow \boldsymbol{\theta}^{(1)}\rightarrow...\rightarrow\boldsymbol{\theta}^{(n-1)}$, which we will refer to as the \textbf{IMP flow}. A schematic illustration of what the IMP flow might look like is given in Fig. \ref{fig:RG IMP} (the axes of which will be explained in Sec. \ref{Results Eigenfunctions}). Just as in the case of RG theory, the IMP flow is determined by the eigenvectors of $\mathcal{T}$, growing or shrinking exponentially by the magnitude of the associated eigenvalues along each direction. 

\subsection{Connection Between the RG and LTH Frameworks}
\label{RG Theory RG and LTH}

Before we formally show that IMP is an RG scheme, we discuss how the standard perspective that has emerged from studying the LTH is related to RG theory. 

In this picture, the success of winning tickets is attributed to the surviving sparse DNN being able to find a ``good'' local minimum in the loss landscape \citep{fra19, fra20}. This may be possible because the training of DNNs via stochastic gradient descent (SGD) is rapidly confined to a low-dimensional subspace \citep{gur18, li2022low}, implying that DNNs only ``feel'' changes to a small number of parameters during much of training. Parameters that are not in this low-dimensional subspace can, therefore, be removed with minimal impact. If a sparse DNN is initialized in this subspace (as late rewinding aims to do), then it may be possible for training to find the same, or related, local minima as the full DNN \citep{evc20, mae21, zhang2021validating}.

In this way, the parameters that do not lie in the low-dimensional subspace are irrelevant, and repeated applications of IMP are expected to remove them. On the other hand, the relevant parameters are those that span this subspace, and their removal changes the local minimum that the sparse DNN converges to. Therefore, certain observable functions of the DNN, such as error, are expected to only be sensitive to the relevant, but not irrelevant, parameters.

If two models share the same low-dimensional subspace, then they should be able to transfer winning tickets. Note that this transferability is highly non-trivial. Indeed, it was only with the development of the LTH, and subsequent experimental work, that this was considered to be a possibility \citep{fra19, meh19, mor19}. It is, in general, difficult to find these subspaces and accurately compare them across experiments. 

RG theory provides us with tools for finding the relevant and irrelevant directions, which can be compared across models. In particular, RG theory says that if two models have the same eigenvectors of $\mathcal{T}$ with eigenvalues $>1$, then they have the same relevant parameters. This means that, once the eigenvectors and eigenvalues are computed, we can potentially know whether winning tickets are transferable between two models, without having to run any additional experiments. Note that the converse, two models having distinct relevant directions, does not directly imply that they cannot transfer winning tickets. However, it does imply that they are differently affected by IMP, suggesting different underlying properties of the models. 

We end by remarking that numerical algorithms for computing RG critical exponents, such as those that make use of Monte Carlo methods \citep{ma1976renormalization, swendsen1979monte, shenker1980monte}, are sensitive to the exact distribution of spins. Therefore, even though the RG theory outlined in this section does not explicitly depend on the amount of coarse-graining used at each application of the RG (although see Eq. \ref{eq:sigma} for how this impacts the $\lambda$), in practice multiple iterations of the RG with small amounts of coarse-graining give the best results. This is similar in spirit to results showing that winning tickets found using multiple rounds of IMP with moderate sparsity outperform tickets found with a single, large sparsity round of IMP (i.e. ``one-shot'' pruning) \citep{fra19}. 

\subsection{IMP as an RG Scheme}
\label{RG Theory RG Scheme}

To make the connection between IMP and RG more precise, we show that IMP fits the definition of an RG scheme. To do this, we consider the projection operator, $\mathcal{P}$, that is associated with the RG operator. In the case of classical spin systems, the projection operator maps the spins, $s_i$, to a coarse-grained spin system, $s'_I$, such that it satisfies 
\begin{equation}
    \label{eq:projection operator hamiltonian}
    \text{Tr}_{\{s_i\}} \mathcal{P}(s_i, s'_I) \exp\left[\mathcal{H}(s_i, \textbf{k})\right] =  \exp\left[\mathcal{H}( s_I', \textbf{k}')\right],
\end{equation}
where Tr$_{\{s_i\}}$ is the trace operator over the values that the $s_i$ can take (e.g. $\pm 1$) \citep{gol05}. For the 2D Ising model, it is standard to take
\begin{equation}
\label{eq:2D Ising projection}
    \mathcal{P}(s_i, s'_I) = \prod_{I} \delta\left[s'_I -  \text{sign}\left(\sum_{j \in I} s_j\right)\right],
\end{equation}
where $\delta$ is the Kronecker delta function and $\text{sign}(\cdot)$ is $+1$ if the argument is positive and $-1$ if it is negative \citep{gol05}. Eq. \ref{eq:2D Ising projection} formally defines the mapping from $s_i \rightarrow s'_I$ as

\begin{equation}
s'_I = \text{sign}\left(\sum_{j \in I} s_j\right) .
\end{equation}

The projection operator is not unique, but must satisfy the following three properties \citep{gol05}: 
\begin{enumerate}
    \item [\ding{182}] $\mathcal{P}(s_i, s'_I) \geq 0$;
    \item [\ding{183}] $\mathcal{P}(s_i, s'_I)$ respects the symmetry of the system;
    \item [\ding{184}] $\displaystyle{\sum_{\{s'_I\}} \mathcal{P}(s_i, s'_I) = 1}$.
\end{enumerate}

To find a projection operator associated with $\mathcal{I}$, we start by finding a mapping between the activations of all the units, $\textbf{a}$, before and after an application of IMP.  This is because $\textbf{a}$, and not $\boldsymbol{\theta}$, is the analogous quantity to $\textbf{s}$ (Table \ref{tab:RG IMP equiv}). Without loss of generality, we can consider the activation of unit $j$ in layer $i$ as being defined by 
\begin{equation}
    \label{eq:activation function}
    a_j^{(i)} = h \left[\sum_{k} g_k(\textbf{a}, \boldsymbol{\theta}) \right],
\end{equation}
where $h$ is the activation function (e.g. ReLu, sigmoid) and the $g_k$ are functions that determine how the different parameters and activations of other units affect $a_j^{(i)}$. For instance, in a feedforward DNN, the impact of the bias of unit $j$ in layer $i$ is given by $g_0 = \theta_j^{(i)}$, and the weighted input from the previous layer is given by $g_1 = \sum_{k = 1}^{N^{(i-1)}} \theta_{jk}^{(i)} a_k^{(i - 1)}$. Here $N^{(i-1)}$ is the number of units in layer $i-1$. 

As discussed above, IMP changes the parameters $\boldsymbol{\theta}$ by the operator $\mathcal{T}$, which is defined by a composition of $\mathcal{M}$ and $\mathcal{F}$. Therefore, the activation of unit $j$ in layer $i$, after applying $\mathcal{I}$, is given by

\begin{equation}
    \label{eq:IMP activation}
    a'_j{^{(i)}} = h \left[\sum_{k} g_k(\textbf{a}, \mathcal{F} \circ \mathcal{M} \boldsymbol{\theta}) \right],
\end{equation}
and thus, the projection operator associated with $\mathcal{I}$ is 
\begin{equation}
\label{eq:IMP projection}
    \mathcal{P}\left(a_j^{(i)}, a'_j{^{(i)}}\right) = \prod_{j = 1}^N \delta \left\{a'_j{^{(i)}} - h \left[\sum_{k} g_k(\textbf{a}, \mathcal{F} \circ \mathcal{M} \boldsymbol{\theta})\right]\right\}.
\end{equation}

We can easily verify that the projection operator defined by Eq. \ref{eq:IMP projection} satisfies all three properties needed for an RG projection operator. First, property \#1 is satisfied, as Eq. \ref{eq:IMP projection} is a product of Kronecker delta functions. Property \#2 is satisfied, as IMP only removes parameters, so the form of $a_j^{(i)}$ given in Eq. \ref{eq:IMP activation} and, thus, the loss function, will remain intact until layer collapse (i.e. when all the weights from one layer to another are removed). Finally, for property \#3 to be satisfied, we can fix the test and training sample ordering for each epoch (as is done when the random seed is fixed). In such a case, both the mask and refining operators in Eq. \ref{eq:IMP projection} are deterministic and $\mathcal{P}\left(a_j^{(i)}, a'_j{^{(i)}}\right)$ will be unique. 

Having found that Eq. \ref{eq:IMP projection} satisfies the three properties necessary for an RG projection operator, we have therefore shown that $\mathcal{I}$ meets the criteria for being an RG operator. To the best of our knowledge, this has not been previously identified and provides new insight into why IMP has found success in discovering winning tickets \citep{fra19, fra20} and an uncovering more general DNN phenomena \citep{frankle2020earlyphase}.

\section{IMP Eigenfunctions}
\label{Results Eigenfunctions}
In order to study the IMP flow, we need to find eigenfunctions of $\mathcal{T}$ from Eq. \ref{eq:IMP eq}. By looking at the eigenvalues corresponding to those eigenfunctions, we can determine the relevant and irrelevant directions. 

For spin systems in statistical physics, the RG flow is studied in the space of coupling constants (Fig. \ref{fig:RG IMP}). Each coupling constant is usually assumed to be the same for all spins, greatly reducing the dimensionality of the underlying space. While most DNNs do not set all parameters of a given type to the same value, we can none-the-less estimate the relative ``influence'' the parameters of a given layer or residual block have on the full DNN, by considering the functions 
\begin{equation}
    \label{eq:layer percent}
    M_i(n) = \sum_{j = 1}^{N^{(i)}} |m_j^{(i)}(n)\cdot \theta_j^{(i)}(n)|\Big/\sum_{k=1}^N |m_k(n) \cdot \theta_k(n)|,
\end{equation}
which describe the percentage of the total remaining parameter magnitude that remains in layer or residual block $i$ after $n$ applications of IMP. For example, $M_1(2) = 0.25$ implies that, after 2 rounds of IMP, layer 1 has $25\%$ of all the remaining weight of the DNN. The numerator in Eq. \ref{eq:layer percent} sums over all $N^{(i)}$ of the parameters restricted to layer or residual block $i$, $\boldsymbol{\theta}^{(i)}$. Multiplying by the corresponding elements of the pruning mask, $\textbf{m}^{(i)} \in \{0, 1\}^{N^{(i)}}$, makes sure that only the non-pruned parameters are considered in the sum. The denominator sums over all $N$ parameters of the DNN, multiplied again by the corresponding mask values, making it equal to the total magnitude of all non-pruned parameters. 

If the $M_i$ are eigenfunctions of the IMP operator, then they will scale exponentially with respect to the number of iterations $\mathcal{T}$ has been applied. That is, $M_i(n + 1) = \mathcal{T} M_i(n) = \lambda_i M_i(n) = \lambda_i^{n + 1} M(0)$, where $\lambda_i$ is the eigenvalue governing its exponential growth/decay. We can find $\lambda_i$ by the simple inversion, $\lambda_i = M_i(n + 1) / M_i(n)$. As long as $M_i(n) \neq 0$, this is well defined and we can approximate $\lambda_i$ by averaging over the computed values from each pair of data points. When we compute the $\lambda_i$ of the computer vision models we study in Sec. \ref{Results}, the standard error of the mean (SEM) is $< 5\%$, supporting the idea that it is appropriate to consider the $M_i$ as eigenfunctions.

Because the degree of coarse-graining at each step (i.e. the amount the DNN gets sparsified, $c \in [0, 1]$, with each round of IMP) affects the magnitude of the eigenvalues, the RG community often considers the quantity $\sigma$,
\begin{equation}
\label{eq:sigma}
    \lambda_i \sim c^{\sigma_i},
\end{equation}
which is invariant to the choice of $c$ \citep{gol05}. Taking $\log_c(\lambda_i)$ gives $\sigma_i$. Because we are interested in comparing across models that prune using different, and possibly even variable, $c$ values, we will report $\sigma_i$ (see Appendix \ref{coarse-graining size} for details on how we determined $c$). Note that, relevant directions, which have $\lambda_i > 1$, have $\sigma_i > 0$, and irrelevant directions, which have $\lambda_i < 1$, have $\sigma_i < 0$.

The $M_i$ are not necessarily the only functions one could consider. Given that the sparsity ratio in each layer or residual block appears to play a crucial role in the success of sparse models \citep{su2020sanity, tanaka2020pruning, frankle2021pruning}, we also evaluated the functions $P_i(n) =  \sum_{j = 1}^{N^{(i)}} m_j^{(i)}(n)/\sum_{k=1}^N m_k(n)$, which correspond to the percent of non-pruned parameters in each layer or residual block. We found that these functions could also be reasonably considered eigenfunctions of $\mathcal{T}$, and the results we got using them to study the IMP flow were consistent with those we got when using the $M_i$ \footnote{Note that this is for trivial reasons for the pre-trained computer vision and natural language processing models we consider in Secs. \ref{Results Computer Vision} and \ref{Results NLP}, as the masks for all the downstream tasks come from the same pre-trained model. However, even in cases when we did not consider pre-trained models, $P_i$ gave qualitatively similar $\sigma_i$ as $M_i$ did.}. Future work will have to determine what set of functions is best to consider.

\section{Results}
\label{Results}
\subsection{Computer Vision Transfer}
\label{Results Computer Vision}

We start by examining the IMP flow of computer vision models, where the LTH and winning ticket transfer have been most thoroughly characterized. The goal of this analysis is to examine models that are known to allow for the successful transfer of winning tickets and determine whether they have the same relevant and irrelevant directions. If they do, this would support the idea that the theory developed in Sec. \ref{RG Theory} properly captures important aspects of sparse machine learning and can be used to identify transferable tickets, without additional experiments.

Because ResNet-50 has four residual blocks, IMP induces a four-dimensional flow in the space of the $M_i$ (Eq. \ref{eq:layer percent}). Example 2D slices of these are plotted in Fig. \ref{fig:IMP_flow} of Appendix \ref{IMP flow} for ResNet-50 trained on CIFAR-10, and CIFAR-100 from random initialization, with 5\% rewind. This data comes from experiments performed by \citet{che21}. 

Computing the $\sigma_i$ (using the procedure described in Sec. \ref{Results Eigenfunctions}) for ResNet-50 trained on CIFAR-10 and CIFAR-100 from random initialization, with 5\% rewind revealed that CIFAR-10 and CIFAR-100 have similar distributions (Table \ref{tab:resnet_lambda}, top row). In particular, both have the first three residual blocks being relevant and the last block as being irrelevant. From the theory discussed in Sec. \ref{RG Theory}, these results imply that both models are in similar universality classes and that winning tickets can be transferred between them. \citet{mor19} indeed confirmed this, finding that winning tickets could be successfully transferred between CIFAR-10 and CIFAR-100 for ResNet-50s. 

\begin{table}[t]
\caption{Mean computed $\sigma_i$, corresponding to the eigenfunctions $M_i$, for ResNet-50 trained from random initialization (top row) and pre-trained (PT) on ImageNet. Results from one experiment each.}
\label{tab:resnet_lambda}
\vskip 0.15in
\begin{center}
\begin{small}
\begin{sc}
\begin{tabular}{ccccc}
\toprule
Task & $\sigma_1$ & $\sigma_2$ & $\sigma_3$ & $\sigma_4$ \\
\midrule
CIFAR-10 & $0.20$ & $0.14$ & $0.07$ & $-0.68$\\
CIFAR-100 & $0.14$ & $0.20$ & $0.04$ & $-0.05$ \\
\hline 
\\
PT CIFAR-10 & $0.15$ & $0.13$ & $0.02$ & $-0.10$\\
PT CIFAR-100 & $0.15$ & $0.22$ & $0.10$ & $-0.16$\\
\bottomrule
\end{tabular}
\end{sc}
\end{small}
\end{center}
\vskip -0.1in
\end{table}

This similar distribution of $\sigma_i$ may be due to the fact that the first few residual blocks learn low level statistics of the data \citep{ney20}, which scale in specific, non-trivial ways for natural images \citep{rud94, sar13}. The particular sensitivity of the first two residual blocks (revealed by their large $\sigma_i$ magnitudes), which are the furthest from the output layer, is in contrast to standard spin systems. In the case of the 2D Ising model, the RG removes long-range interactions (i.e. couplings between next next-nearest neighbors, etc.), which are typically assumed, for physical reasons, to be weak. Systems with relevant long-range interactions have been found to have unique properties, such as the existence of a phase transition in 1D \citep{dys69}.

Recent work has found that using DNN parameters from models that have been pre-trained on complex tasks allows for substantial transfer \citep{che20, che21}. We therefore examined the effect of pre-training using ImageNet \citep{ImageNet_transfer}. We find that, while the winning tickets found using the pre-trained parameter values identify the same relevant and irrelevant residual blocks as with the randomly initialized case, CIFAR-10's $\sigma_4$ is pushed considerably closer to CIFAR-100's $\sigma_4$ (Table \ref{tab:resnet_lambda}, bottom row). This suggests that the success transferring winning tickets has found after pre-training comes, in part, from constraining the IMP flow, making the $\sigma_i$ more similar across downstream tasks. Note that this is in agreement with intuition that pre-training induces biases. We find similar results when computing the $\sigma_i$ of ResNet-50, pre-trained on ImageNet, with no task specific refining (Table \ref{tab:ImageNet_lambda} of Appendix \ref{Additional CV results}), although differences in experimental set-up prevent direct comparisons between $\sigma_i$ values.

\subsection{Natural Language Processing Transfer}
\label{Results NLP}

Winning tickets and their transfer have also been studied in the context of natural language processing (NLP) models \citep{yu2020, che20, prasanna2020bert}. We computed the $\sigma_i$ on tickets that were found by applying IMP to pre-trained BERT models on ten downstream NLP tasks \citep{rajpurkar2016squad, wang2018glue}, which were known to allow for ticket transfer. Data comes from experiments performed by \citet{che20}. 

Similar to the computer vision transfer results of Sec. \ref{Results Computer Vision}, we find that tasks that have transferable winning tickets have very similar $\sigma_i$ (Table \ref{tab:bert_lambda}). That the $\sigma_i$, for different tasks, so nearly match each other provides further support for the idea that pre-training confines the IMP flow.

However, there is an important difference between the ResNet-50 and BERT results. Namely, all the BERT $\sigma_i$ (other than $\sigma_1$) are closely clustered around $0$, being one to two orders of magnitude smaller than those of ResNet-50. This suggests that, unlike pre-trained ResNet-50, pre-trained BERT has a considerably more static IMP flow, possibly being near a fixed point.

Previous work on transformers applied to computer vision tasks have shown that self-attention heads in consecutive layers can become extremely similar \citep{touvron2021going, gong2021vision, zhou2021deepvit}. If BERT layers are likewise learning highly overlapping features, then it would be natural to find it at an IMP fixed point. Future work should investigate how methods that have been developed to break this self-similarity \citep{touvron2021going, gong2021vision, zhou2021deepvit} affect the IMP eigenvalues and the transferability of tickets between tasks. 

\begin{table*}[]
\caption{Mean computed $\sigma_i$, corresponding to the eigenfunctions $M_i$, for pre-trained BERT on ten downstream tasks NLP tasks \citep{rajpurkar2016squad, wang2018glue}. Results from one experiment each.}
\label{tab:bert_lambda}
\vspace{-3mm}
\begin{center}
\begin{small}
\begin{sc}
\begin{tabular}{cccccccccccccc}
\toprule
Task & $\sigma_1$ & $\sigma_2$ & $\sigma_3$ & $\sigma_4$ & $\sigma_5$ & $\sigma_6$ & $\sigma_7$ & $\sigma_8$ & $\sigma_9$ & $\sigma_{10}$ & $\sigma_{11}$ & $\sigma_{12}$ \\
\midrule
MNLI & $-0.054$ & $-0.005$ & $-0.001$ & $0.002$ & $0.017$ & $0.018$ & $0.013$ & $-0.010$ & $-0.004$ & $0.013$ & $0.007$ & $0.004$\\
QQP & $-0.056$ & $-0.004$ & $-0.001$ & $0.001$ & $0.017$ & $0.018$ & $0.013$ & $-0.010$ & $-0.003$ & $0.014$ & $0.007$ & $0.005$\\
STS-B & $-0.058$ & $-0.004$ & $-0.002$ & $0.001$ & $0.016$ & $0.018$ & $0.013$ & $-0.010$ & $-0.003$ & $0.014$ & $0.008$ & $0.006$\\
WNLI & $-0.058$ & $-0.004$ & $-0.002$ & $0.001$ & $0.016$ & $0.018$ & $0.013$ & $-0.010$ & $-0.003$ & $0.015$ & $0.008$ & $0.006$\\
QNLI & $-0.057$ & $-0.004$ & $-0.001$ & $0.001$ & $0.017$ & $0.018$ & $0.013$ & $-0.010$ & $-0.003$ & $0.014$ & $0.007$ & $0.005$\\
MRPC & $-0.058$ & $-0.004$ & $-0.002$ & $0.001$ & $0.016$ & $0.018$ & $0.013$ & $-0.010$ & $-0.002$ & $0.015$ & $0.008$ & $0.006$\\
RTE & $-0.058$ & $-0.004$ & $-0.001$ & $0.001$ & $0.016$ & $0.018$ & $0.014$ & $-0.010$ & $-0.003$ & $0.014$ & $0.008$ & $0.006$\\
SST-2 & $-0.058$ & $-0.004$ & $-0.001$ & $0.001$ & $0.017$ & $0.018$ & $0.013$ & $-0.010$ & $-0.003$ & $0.014$ & $0.007$ & $0.006$\\
CoLA & $-0.058$ & $-0.004$ & $-0.002$ & $0.001$ & $0.016$ & $0.018$ & $0.013$ & $-0.010$ & $-0.002$ & $0.014$ & $0.007$ & $0.006$ \\ 
SQuAD v1.1 & $-0.052$ & $-0.005$ & $-0.001$ & $0.002$ & $0.017$ & $0.018$ & $0.014$ & $-0.009$ & $-0.003$ & $0.012$ & $0.005$ & $0.004$ \\
\bottomrule
\end{tabular}
\end{sc}
\end{small}
\end{center}
\vskip -0.1in
\end{table*}

\subsection{Elastic Lottery Ticket Hypothesis}
\label{Results ELTH}

Recent work on the ``Elastic Lottery Ticket Hypothesis'' (E-LTH) has extended the notion of tranferability by finding it possible to transform winning tickets found on one model to another with a different architecture \citep{chen2021ELTH}. In particular, it was found that, for architectures in the same family (e.g. ResNets), trained on the same task, it was possible to either squeeze (by removing residual blocks) or stretch (by replicating residual blocks) winning tickets so that they could be transferred. 

Three results from this study are especially noteworthy \citep{chen2021ELTH}: 
\begin{enumerate}
    \item [\ding{182}]  The smallest ResNets considered (ResNet-14 and ResNet-20) had the weakest transferability properties;
    \item [\ding{183}] The more unique residual blocks replicated, the better the performance;
    \item [\ding{184}] Removing the later residual blocks, in the case of shrinking, or replicating the earlier residual blocks, in the case of stretching, led to the best results.
\end{enumerate}
While preliminary hypotheses on the origin of these results were developed by \citet{chen2021ELTH}, involving dynamical systems and unrolled estimation, detailed understanding was left to future work. We examined whether the RG framework developed here could provide further insight into these results. 

We computed the $\sigma_i$ for each ``normal'' residual block of ResNet-14/20/32/44/56, trained on CIFAR-10 [data from \citet{chen2021ELTH}]. As discussed in the original paper, the down-sampling residual block in each stage was not transformed between winning tickets, so they were not considered. The $\sigma_i$ for the residual blocks in the first stage, of all the architectures considered, are presented in Table \ref{tab:ELTH lambda}. 

\begin{table}[]
    \centering
    \caption{Mean computed $\sigma_i$, corresponding to the eigenfunctions $M_i$, for the first stage of various ResNet architectures. Results are mean across three experiments.}
    \begin{tabular}{cc}
        \\
        \small Architecture & \small $\sigma_i$ \\
         \hline
        \small ResNet-14 & \small $0.23$ \\ 
        \small ResNet-20 & \small $0.23$, $0.26$ \\
        \small ResNet-32 & \small $-0.04$, $0.19$, $0.18$, $0.26$ \\ 
        \small ResNet-44 & \small $-0.03$, $-0.00$, $0.20$, $0.19$, $0.19$, $0.23$ \\
        \small ResNet-56 & \small $0.06$, $-0.10$, $0.25$, $0.20$, $0.18$, $0.21$, $0.11$, $0.24$\\
\end{tabular}
    \label{tab:ELTH lambda}
\end{table}

We find that, while the smaller ResNets (ResNet-14 and ResNet-20) have only relevant residual blocks, the larger ResNets have at least one non-relevant block, and have a broader distribution of positive $\sigma_i$ values. Therefore, using ResNet-14 or ResNet-20 as the source ticket necessarily leads to the transformed ticket having residual blocks with larger relative weighting than the target ticket (because of the larger $\sigma_i$). This is likely related to the idea that smaller ResNet models are not sufficiently expressive. In addition, we find that replicating few unique residual blocks (i.e. replicating only block 1 or blocks 1 and 2) can lead to an over representation of either relevant or irrelevant blocks. This again leads to a mismatch in the relative weighting of each residual block between the transformed and target tickets, likely making transfer less effective.

Finally, we find that most of the residual blocks with $\sigma_i < 0$ are in the early part of the first stage. Therefore, removing the first several blocks when shrinking a winning ticket (e.g. dropping the first four blocks to go from ResNet-56 to ResNet-32), or replicating the last several blocks when stretching a winning ticket (e.g. adding the last two blocks to go from ResNet-32 to ResNet-44), will again result in the transformed source ticket having a different structure than the target ticket.

Note that for the third stage, the nature of the distribution of $\sigma_i$ is different (Table \ref{tab:ELTH lambda stage 3} of Appendix \ref{Additional ELTH}). In particular, Stage 3 has only marginal and non-relevant blocks. However, there is consistent evidence that the distributions of $\sigma_i$ in the second and third stages (Tables \ref{tab:ELTH lambda stage 2} and \ref{tab:ELTH lambda stage 3} of Appendix \ref{Additional ELTH}) are similarly related to the three results from the original E-LTH paper \citep{chen2021ELTH} we highlighted. 

\section{Discussion}
Inspired by similarities between the current state of sparse machine learning and the state of statistical physics in the early-to-mid--20$^{\text{th}}$ century, we found that iterative magnitude pruning (IMP), the principal method used to discover winning tickets, is a renormalization group (RG) scheme. Given that the development of the RG led to a first principled understanding of universal behavior near phase transitions, as well as a way in which to characterize materials by such behavior, we reasoned that viewing IMP from an RG perspective may prove useful when studying the universality of winning tickets \citep{desai2019shifts, meh19, mor19, soe19, che20, goh20, che21, sab21} and interpreting the general success IMP has found as a tool for interrogating DNNs \citep{frankle2020earlyphase}. 

By viewing IMP as inducing a flow in DNN parameter space (Eq. \ref{eq:IMP flow} and Fig. \ref{fig:IMP_flow} of Appendix \ref{IMP flow}), we showed that ResNet-50 models have trajectories with common properties across different computer vision tasks (Table \ref{tab:resnet_lambda}). Winning tickets are known to be transferable between these models \citep{mor19}, suggesting that the theory developed in Sec. \ref{RG Theory} captures important properties of sparse machine learning models and may be useful in identifying transferability, without requiring additional experiments.

Winning tickets found on ResNet-50 models pre-trained on ImageNet led to similar identification of relevant and irrelevant residual blocks. However, the values of $\sigma_4$, which had differed most significantly in the case of no pre-training, were substantially more alike across tasks (Table \ref{tab:resnet_lambda}). This suggests that pre-training constrains the IMP flow, enabling better transfer. Studying winning tickets found on natural language processing tasks, using pre-trained BERT models, again identified $\sigma_i$ distributions that were very similar across downstream tasks (Table \ref{tab:bert_lambda}). Unlike the $\sigma_i$ of the pre-trained ResNet-50, the $\sigma_i$ of the pre-trained BERT were tightly clustered around $0$, suggesting that the models are near a fixed point of the IMP flow. This is consistent with work that has found that vision transformers can develop self-attention heads that are nearly identical across layers \citep{touvron2021going, gong2021vision, zhou2021deepvit}. 

Finally, we applied the RG framework to recent work that extended the notion of winning ticket universality by transforming tickets between different architectures \citep{chen2021ELTH}. We found that the distributions of $\sigma_i$ in the smallest ResNet architectures were more uniform than the distributions of the largest ResNet architectures. Therefore, using them as source tickets leads to a mismatch in structure with target tickets. This is in-line with the observation that the smallest ResNet architectures had the worst transfer properties. We additionally found that the $\sigma_i$ values offered interpretations of other experimental results found by \citet{chen2021ELTH}.

\subsection{Future Directions}
A wealth of literature  has been developed around the RG, including a number of numerical and theoretical tools that go beyond what we have used here. Given the connection between IMP and the RG, made explicit for the first time in this work, we believe that there is considerable potential for collaboration between the two fields. Beyond the ideas already noted, possible directions of future study include:
\begin{itemize}
    \item Bringing the RG framework to other, non-IMP sparsificiation methods;
    \item Computing the $\sigma_i$ associated with models that are used outside of computer vision and natural language processing, such as in the context of reinforcement learning and lifelong learning, where the LTH has been studied \citep{yu2020, chen2021_lifelong};
    \item Computing and classifying systems by their critical exponents (e.g. $\gamma$ in Eq. \ref{eq:pruning scaling law}). We attempted to do this, but we were limited by having only a few independent seeds and a scaling function with multiple free parameters \citep{ros21} (see Appendix \ref{Critical Exponents}). Using more advanced methods, such as finite scaling \citep{gol05}, and increasing the number of independent seeds may allow for better results;
    \item Developing novel pruning methods. The RG has a complimentary, momentum space perspective, to the real space perspective presented in this manuscript (Appendix \ref{RG theory in physics}). This framework allows for coarse-graining at continuous, instead of discrete, scale. Whether this is possible in the context of DNN pruning is an exciting question;
    \item Identifying, via the RG framework, the minimal density a winning ticket can have and still be transferred to a given task.
\end{itemize}

\section*{Acknowledgements}

We would like to thank Jonathan Rosenfeld for his advice on performing the fits described by Eq. \ref{eq:gamma eq} (Appendix \ref{Critical Exponents}).

W.T.R. is partially supported by a UC Chancellor’s Fellowship. T.C. is supported by an IBM PhD
fellowship. Z.W. is supported by a US Army Research Office Young Investigator Award (W911NF2010240). A.S.D.’s research is supported by the EPSRC Centre for Doctoral Training in Mathematics of Random Systems: Analysis, Modelling and Simulation (EP/S023925/1). A.S.D. is funded by the President’s PhD Scholarships at Imperial College London. 

\bibliography{main}
\bibliographystyle{icml2022}

\newpage
\appendix
\onecolumn

\section{Brief Background on Renormalization Group Theory in Physics}
\label{RG theory in physics}
Phase transitions have long fascinated physicists. These transitions can be between phases of matter (e.g. liquid, solid, gas), but more generally refer to regimes where a continuous, systematic change in one parameter (called the control parameter) leads to a divergence in another parameter (called the order parameter), or its derivative. Examples of this include the ferromagnetic phase transition, where a metal goes from being non-magnetic to magnetic as it is cooled (see Fig. \ref{fig:RG IMP}). 

The point at which the divergence occurs is called the critical point. Specific phenomena occur near critical points, such as power-law scaling of certain observables. As discussed in the main text, $m \sim (t_C - t)^{-\beta}$, where $m$ is the magnetization, $t_C$ is the critical temperature, and $t$ is the temperature of the system (Eq. \ref{eq:magnetization scaling law}). The exponent present in the power-law scaling is called a critical exponent. A major discovery of 20$^{\text{th}}$ century physics was that systems with seemingly different underlying physics exhibited power-law scaling with \textit{the same} critical exponents. For example, the liquid gas phase transition also has an order parameter, the difference in density between gas and liquid, which scales as $|\rho_+ - \rho_-| \sim |t_C - t|^{-\beta}$. The experimentally measured values of $\beta$, in the case of the ferromagnetic transition, are nearly identical to experimental values of $\beta$, in the case of the liquid gas transition \citep{gol05}. 

Various methods were proposed to understand phases transitions, such as mean field theory and Landau theory, however none could properly compute the critical exponents for two and three dimensional systems. The introduction of the renormalization group (RG) \citep{wil71A, wil71B, wil75} solved this problem by providing an explicit, albeit challenging, way in which to compute critical exponents and understand their universality. 

The formulation of the RG considered in the main text (i.e. the ``block spin'' or ``real space'' RG) considers iterative transformations of physical space (i.e. ``coarse-graining'') as a kind of dynamical system, whose linearization around fixed points of the RG flow gives the associated critical exponents. Note that this is an approximation to the true nonlinear nature of the RG operator $\mathcal{R}$, and corrections to these computed critical exponents can be obtained. A general framework with which to do this via normal form theory has recently been proposed \citep{raju2019normalform}. Beyond this, \citet{red20} identified the RG as a Koopman operator \citep{mezic2005spectral} (i.e. an infinite dimensional linear operator in function space, directly associated with the nonlinear dynamics in state space). If a suitably invariant subspace could be found, then the RG eigenvalues could in principle be computed as eigenvalues of the Koopman operator, without needing to perform any linearization. 

In addition to providing insight into the nature of critical exponents and universality classes, the real space RG has been used to analytically and numerically compute them. For example, in the former case, RG methods have been used to study many-body localization \citep{vosk2013manybody}, percolation \cite{reynolds1977real}, the ``route to chaos'' \citep{feigenbaum1982quasiperiodicity}, and small-world networks \citep{newman1999renormalization}. In the latter case, Monte Carlo methods have been adapted to the RG \citep{ma1976renormalization, swendsen1979monte, shenker1980monte}, making it possible to study complicated spin systems, like the Ising spin glass \citep{wang1988monte, katzgraber2006universality, jorg2008universality}. 

In addition to the real space RG, there is a related framework for using the RG, called the momentum space RG. As the name suggests, instead of removing degrees of freedom in physical space, this approach removes Fourier modes of the system with large momentum, and thus, high energy. Because of the relationship between energy and wavelength, this in effect integrates out short wavelength (i.e. local) degrees of freedom, again coarse-graining the system in physical space. While the end result is similar, the threshold for what constitutes as ``large momentum'' can be continuously varied. This allows for a more refined coarse-graining that is done using the real space RG. The momentum space RG plays a major role in quantum field theories, and has also been used to compute critical exponents in dimensions less than 4 \citep{wilson1972critical} and extend principal component analysis \citep{bradde2017pca}. 

We end by noting that the distinction between relevant and irrelevant interactions have been found in many systems, including biological models and feedforward neural networks \cite{gutenkunst2007universally, machta2013parameter}.

\section{Determining $c$}
\label{coarse-graining size}

In the standard RG theory, the amount of coarse-graining that occurs during each application of the RG operator is given by $b$. This corresponds to the size, in each dimension, of the neighborhood over which the spins get amalgamated. For instance, in the case of the two-dimensional (2D) Ising model example presented in the left most column of Fig. \ref{fig:RG IMP}, $b = 2$. Note that one way to interpret $b$ is that it corresponds to how many of the spins in the original system get turned into one spin in the coarse-grained system (for each dimension).

Using this idea, we can compute the amount of coarse-graining that occurs after each round of IMP by considering how the number of weights a given unit has (on average) changes. If a DNN is sparsefied by an amount $x\in (0, 1)$ each round of IMP, then, for every set of $1/(1-x)$ weights in the original DNN, there will be $1$ weight in the sparsified DNN. Therefore, we should define the amount of coarse-graining, $c$, to be $c = 1/(1 - x)$.

In the case of the computer vision experiments (Secs. \ref{Results Computer Vision} and \ref{Results ELTH}), the models were sparsified $20\%$ each round. Therefore, $ x= 0.2$ and $c = 1/(1 - 0.2) = 1/0.8 = 1.25$. Intuitively, this corresponds to the fact that for every 10 weights in the model that is to be pruned, only 8 would remain after a round of IMP. 

In the case of the natural language processing experiments (Sec. \ref{Results NLP}), $10\%$ of the original number of weights are pruned each round. This means that, $x$ changes as a function of the number of rounds the DNN has been pruned. Initially, $x(0) = 0.1$ and $c = 1/(1-0.1) = 1.\bar{1}$. As $x(n)$ increases with increasing $n$, so too does $c$ increase. To compute $c$ at IMP round $n$, we again use $c(n) = 1/(1 - x(n))$, where $x(n) = 1/(10 - n)$. Thus, when $n = 2$, $c(2) = 1/[1 - 1/(10 - 2)] = 8/7 = 1.14...$.

\section{IMP Flow}
\label{IMP flow}

\begin{figure}[h!]
    \centering
    \includegraphics[width = 0.75\textwidth]{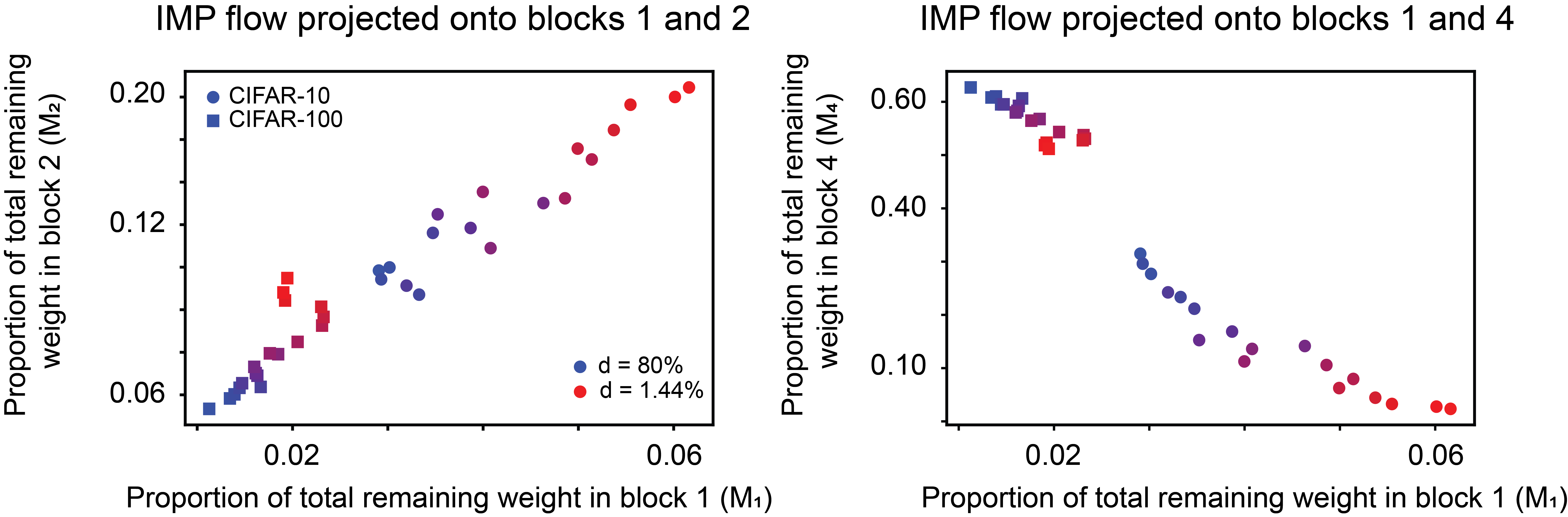}
    \caption{\textbf{IMP flow.} Two-dimensional projections of the IMP flow for ResNet-50 trained from random initialization, with $5\%$ rewind, applied to two different tasks: CIFAR-10 (circles), and CIFAR-100 (squares). Blue corresponds to a single application of IMP (density equal to $80\%$) and red corresponds to 17 applications of IMP (density equal to $1.44\%$).}
    \label{fig:IMP_flow}
\end{figure}

As discussed in the main text, we chose to analyze the IMP flow by considering its action on the functions $M_i(n)$ (Eq. \ref{eq:layer percent}). These describe the proportion of parameter magnitude remaining in layer or residual block $i$ after $n$ rounds of IMP. We found that these functions evolved in an exponential manner, described by the constant $\lambda_i$. This made us consider them to be eigenfunctions of $\mathcal{T}$ (Eq. \ref{eq:IMP eq}), with $\lambda_i$ being eigenvalues that determine whether residual block is relevant, irrelevant, or marginal (defined in the RG sense -- see Sec. \ref{RG Theory}). In Fig. \ref{fig:IMP_flow}, we visualize the flow projected onto two of the eigenfunctions. While the curves do not start from the same point (implying distinct distributions of weights for different tasks), they evolve at a similar rate (determined by the $\lambda_i$ -- see Table \ref{tab:resnet_lambda} for the reported values of $\log_c \lambda_i$). 

\section{Additional Computer Vision Transfer Results}
\label{Additional CV results}

In addition to the computer vision transfer results that we present in the main text (Table \ref{tab:resnet_lambda}), we also computed the $\sigma_i$ for a ResNet-50 pre-trained on ImageNet with no fine-tuning (Table \ref{tab:ImageNet_lambda}). The exact way in which IMP was applied was slightly different from the experiments presented in the main text (namely, all layers of each residual block could be pruned, including the first one). Given these differences, we decided to report the results separately from those of the main text. However, we reach a similar conclusion as we did for the other ResNet-50 experiments (Table \ref{tab:resnet_lambda}). Namely that the first three residual blocks are relevant, and the last one is irrelevant.

\begin{table}[t]
\caption{Mean computed $\sigma_i$, corresponding to the eigenfunctions $M_i$, for ResNet-50 pre-trained on ImageNet with no fine-tuning. Results from one experiment.}
\label{tab:ImageNet_lambda}
\vskip 0.15in
\begin{center}
\begin{sc}
\begin{tabular}{ccccc}
\toprule
Network & $\sigma_1$ & $\sigma_2$ & $\sigma_3$ & $\sigma_4$ \\
\midrule
Pre-trained ImageNet & $0.45$ & $0.30$ & $0.08$ & $-0.24$\\
\end{tabular}
\end{sc}
\end{center}
\end{table}

\section{Additional E-LTH Results}
\label{Additional ELTH}
In the main text on the experimental results surrounding the ``Elastic Lottery Ticket Hypothesis'' (E-LTH), we reported the $\sigma_i$ corresponding to the eigenfunctions of the residual blocks in the first stage. The models considered by \citet{chen2021ELTH} had two other stages, whose $\sigma_i$ we report here in Tables \ref{tab:ELTH lambda stage 2} and \ref{tab:ELTH lambda stage 3}.

While the distributions of $\sigma_i$ are different than that of Stage 1, we again find that ResNet-14 and ResNet-20 have residual blocks that are more uniform in their sensitivity to pruning (i.e. have $\sigma_i$ with more similar values) than the larger models.

\begin{table}[h]
    \centering
    \caption{Computed $\sigma_i$, corresponding to the eigenfunctions $M_i$, for the second stage of various ResNet architectures. Results are mean across three experiments.}
    \label{tab:ELTH lambda stage 2}
    \begin{tabular}{cc}
        \\
        ResNet architecture & $\sigma_i$ \\
         \hline
        ResNet-14 & $0.10$\\ 
        ResNet-20 & $0.08$, $0.07$\\
        ResNet-32 & $0.11$, $0.11$, $0.04$, $-0.01$ \\ 
        ResNet-44 & $0.15$, $0.11$, $0.10$, $0.09$, $0.02$, $0.04$\\
        ResNet-56 & $0.10$, $0.06$, $0.06$, $0.06$, $0.02$, $-0.01$, $-0.06$, $0.03$\\
\end{tabular}
\end{table}

\begin{table}[h]
    \centering
    \caption{Computed $\sigma_i$, corresponding to the eigenfunctions $M_i$, for the third stage of various ResNet architectures. Results are mean across three experiments.}
    \label{tab:ELTH lambda stage 3}
    \begin{tabular}{cc}
        \\
        ResNet architecture & $\sigma_i$ \\
         \hline
        ResNet-14 & $-0.21$ \\ 
        ResNet-20 & $-0.16$, $-0.20$\\
        ResNet-32 & $-0.06$, $-0.09$, $-0.13$, $-0.16$ \\ 
        ResNet-44 & $-0.09$, $-0.13$, $-0.16$, $-0.20$, $-0.15$, $-0.08$\\
        ResNet-56 & $-0.11$, $-0.12$, $-0.13$, $-0.15$, $-0.18$, $-0.12$, $-0.04$, $-0.03$\\
\end{tabular}
\end{table}

\section{Critical Exponents}
\label{Critical Exponents}

Following the recent work on scaling laws for IMP developed by \citet{ros21}, we fit the error, $\epsilon$, defined as $100\%$ minus the top-1 accuracy, as a function of density, $d$, defined as percent of weights remaining, via the functional form 

\begin{equation}
    \label{eq:gamma eq}
\begin{split}
    \hat{\epsilon}(d, \epsilon_{np}, \epsilon^{\uparrow}, \gamma, p) &= \epsilon_{np} \left|\left|\frac{d - ip\left( \frac{\epsilon^{\uparrow}}{\epsilon_{np}}\right)^\frac{1}{\gamma}} {d - ip}\right|\right|^{\gamma} \\
    &= \epsilon_{np} \left[\frac{\left(d^2 + p^2 (\frac{\epsilon^{\uparrow}}{\epsilon_{np}})^{2/\gamma}\right)}{\left(d^2 + p^2 \right)}\right]^{\gamma/2},
\end{split}
\end{equation}
where $i = \sqrt{-1}$. Here $\epsilon_{np}$ is interpreted as the error associated with not pruning, $\epsilon^{\uparrow}$ as the asymptotic error upon maximal pruning, $\gamma$ as the sensitivity of the combination of network architecture, task, activation function, and optimizer to pruning, and $p$ as controlling how the transition from no change in error to power-law scaling takes place. Importantly, $\gamma$ can be viewed as a critical exponent under the RG framework (Eq. \ref{eq:pruning scaling law}). Because systems in the same universality class have the same critical exponents, we examined whether we could find evidence of universality in the computed value of $\gamma$ for various different computer vision models. 

We found the fits to be sensitive to $\epsilon_{np}$ and, in some cases, there was not a clear single value for $\epsilon_{np}$. Therefore, we included $\epsilon_{np}$ as a free parameter (which \citet{ros21} did not do), but we tightly bounded the value so as to minimize instability in adding another free parameter. To numerically compute the fits, we used scipy's curve fitting function: scipy.optimize.curve\_fit \citep{scipy}.

The computed values of $\gamma$ for ResNet-50 evaluated on CIFAR-10, CIFAR-100, and SVHN are presented in Table \ref{tab:vision gamma}. The effect of pre-training using various methods is also given. The $\gamma$ values vary, sometimes considerably, across tasks and method of pre-training (e.g. CIFAR-10 and SVHN pre-trained via simCLR). However, we note that the data came from only two or three seeds, making the fits susceptible to noise and making the quantification of error in the computed $\gamma$ difficult. In addition, the fitting function of Eq. \ref{eq:gamma eq} has multiple free parameters (including an additional one we added), again making the fit susceptible to noise. 

\begin{table}
    \centering
    \caption{Critical exponent, $\gamma$, for ResNet-50 with either no pre-training and 5\% rewinding, or pre-training using ImageNet or simCLR. Data from \citet{che21}. Results are from fitting Eq. \ref{eq:gamma eq} to mean of two or three seeds.}
    
    \begin{tabular}{c|c|c|c|c}
    \\
         Data-set & No pre-train &  ImageNet &  simCLR & MoCo \\
         \hline
         CIFAR-10 & $\gamma$ = -0.14 & $\gamma$ = -0.10 & $\gamma$ = -0.16 & $\gamma$ = -0.21\\
         CIFAR-100 & $\gamma$ = -0.11 & $\gamma$ = -0.15 & $\gamma$ = -0.06 & $\gamma$ = -0.24\\
         SVHN & $\gamma$ = -0.07 & $\gamma$ = -0.06 & $\gamma$ = -0.03 & $\gamma$ = -0.29 \\
         \hline 
         ImageNet & -- & $\gamma$ = -0.42 & -- & --\\
         simCLR & -- & -- & $\gamma$ = -1.01 & --\\
    \end{tabular}
    \label{tab:vision gamma}
\end{table}

\begin{figure}[t]
    \centering
    \includegraphics[width = 0.45\textwidth]{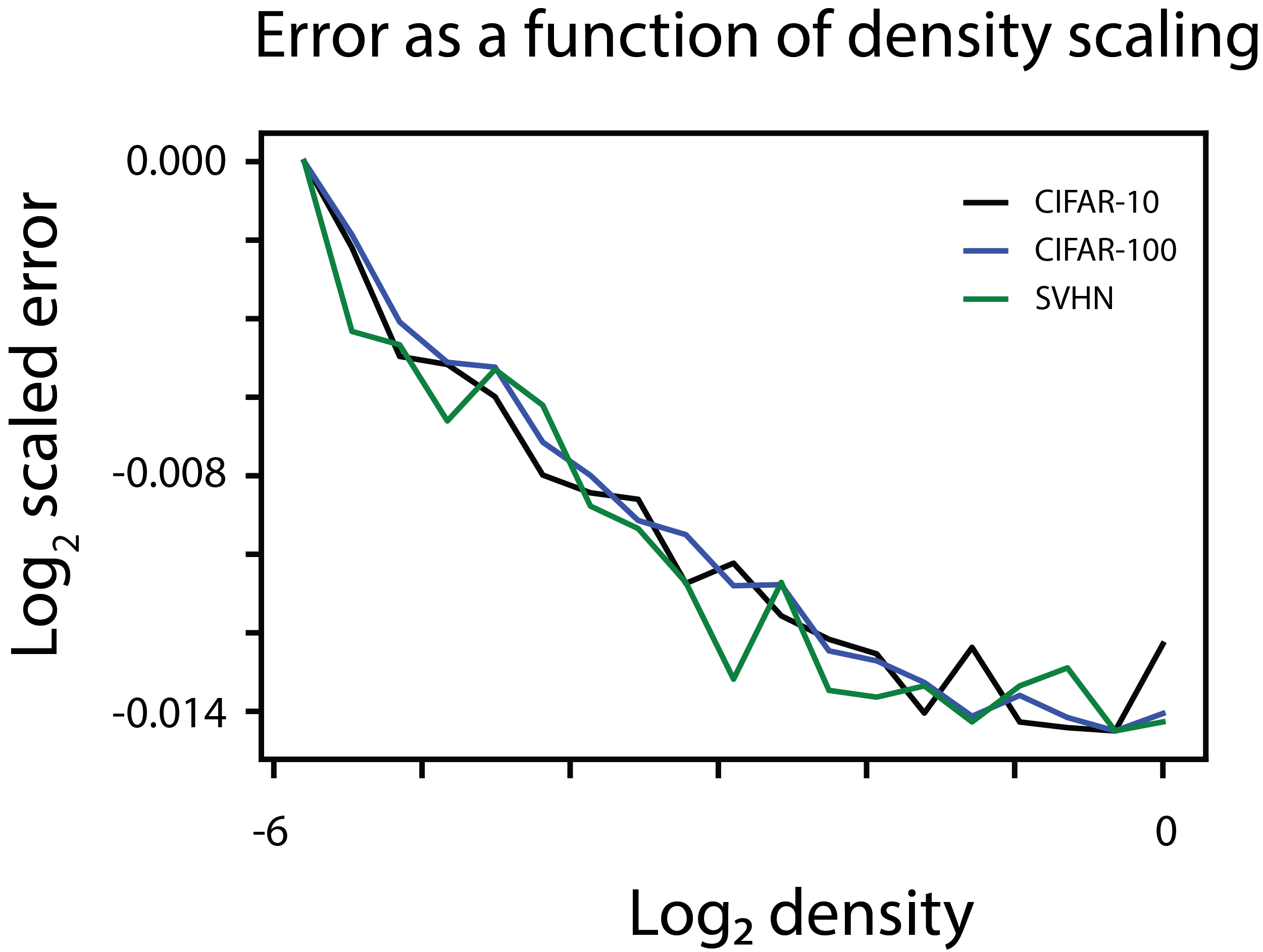}
    \caption{\textbf{Error, as a function of density, scaling of CIFAR-10/100 and SVHN.} Overlaying the error, as a function of density, curves for ResNet-50, trained from random initialization, on CIFAR-10/100 or SVHN, with 5\% rewind. The overlaid curves shows similar, albeit noisy, behavior.} 
    \label{fig:LT_rewind_overlay}
\end{figure}

Taking these possible complications into account, we overlaid the different error as a function of density scaling curves (an example of which is given in Fig. \ref{fig:LT_rewind_overlay}). We found that they qualitatively matched each other more than the computed $\gamma$ may have made it seem. We imagine that trying different fitting procedures [especially making use of methods developed in the context of statistical physics, such as finite scaling \citep{gol05}], as well as using more independent random seeds, will enable more accurate and robust approximations of $\gamma$. 

\end{document}